\documentclass{article}
\usepackage{spconf,amsmath,graphicx}
\usepackage[section]{placeins}
\usepackage{enumitem}
\usepackage{amsmath}      
\usepackage{amssymb}      
\usepackage{mathtools}    
\usepackage{bm}           
\usepackage{wrapfig}
\usepackage[font=small,labelfont=bf]{caption}
\usepackage[T1]{fontenc}
\usepackage[utf8]{inputenc}
\usepackage{siunitx, booktabs, amsmath}
\usepackage{graphicx}     
\usepackage{caption}      
\usepackage{subcaption}   

\usepackage{booktabs}     
\usepackage{multirow}     

\usepackage[ruled,vlined]{algorithm2e}

\usepackage{hyperref}     
\usepackage{cleveref}     

\usepackage{xcolor}

\usepackage{enumitem}     

\title{Learning glioblastoma tumor heterogeneity using brain inspired topological neural networks}
\name{Author(s) Name(s)\thanks{.}}
\address{Author Affiliation(s)}
\address{Author Affiliation(s)}
\name{Ankita Paul$^{1}$, Wenyi Wang$^{1}$}
\address{$^{1}$The University of Texas MD Anderson Cancer Center\\Department of Bioinformatics and Computational Biology\\Houston, TX, USA}

\begin{document}
%
\maketitle
\begin{abstract}
Accurate prognosis for Glioblastoma (GBM) using deep learning (DL) is hindered by extreme spatial and structural heterogeneity. Moreover, inconsistent MRI acquisition protocols across institutions hinder generalizability of models. Conventional transformer and DL pipelines often fail to capture the multi-scale morphological diversity such as fragmented necrotic cores, infiltrating margins, and disjoint enhancing components—leading to scanner-specific artifacts and poor cross-site prognosis. We propose \textbf{TopoGBM}, a learning framework designed to capture heterogeneity-preserved, scanner-robust representations from multi-parametric 3D MRI. Central to our approach is a 3D convolutional autoencoder regularized by a topological regularization that preserves the complex, non-Euclidean invariants of the tumor’s manifold within a compressed latent space. By enforcing these topological priors, TopoGBM explicitly models the high-variance structural signatures characteristic of aggressive GBM. Evaluated across heterogeneous cohorts (UPENN, UCSF, RHUH) and external validation on TCGA, TopoGBM achieves better performance (C-index 0.67 test, 0.58 validation), outperforming baselines that degrade under domain shift. Mechanistic interpretability analysis reveals that reconstruction residuals are highly localized to pathologically heterogeneous zones, with tumor-restricted and healthy tissue error significantly low (Test: 0.03, Validation: 0.09). Furthermore, occlusion-based attribution localizes approximately 50\% of the prognostic signal to the tumor and the diverse peri-tumoral microenvironment advocating clinical reliability of the unsupervised learning method. Our findings demonstrate that incorporating topological priors enables the learning of morphology-faithful embeddings that capture tumor heterogeneity while maintaining cross-institutional robustness.
\end{abstract}
\section{Introduction}
\label{sec:intro}
Glioblastoma (GBM) is a leading cause of mortality in brain cancer with limited treatment options \cite{schaff2023glioblastoma,tamimi2017epidemiology}. Determining the aggressiveness of the cancer from MRI is critical for designating therapy roadmap by clinicians. At present radiologist determined tumor aggressiveness progression from MRI is the only clinically reliable way to inform clinicians on prognosis. Supervised Deep learning (DL) has shown promising direction to extract features from MRI for early detection and onset aggressiveness determination for cancers. Yet this roadmap of therapy for aggressive GBM is frequently unreliable and ineffective in most cases \cite{omuro2013glioblastoma, cerono2024clinical,gomaa2024comprehensive} because of such models are black box in nature, are heavily overfit on their training cohorts which make them unexplainable and not generalizable across cohorts. Moreover, manual feature annotation is required for reliability. The critical gap still remains in learning biologically meaningful features from tumor microenvironment of the MRI, to make clinically relevant decisions. Especially for GBM, extracting tumor periphery, tumor core, tumor holes and tumor microenvironment irregularities from MRI is very challenging with existing unsupervised or supervised approaches. Hence, clinical decision making rarely relies on artificial intelligence (AI) driven feature extraction and modeling for GBM disease.\\
\textbf{Related Works: }DL has transformed neuro-oncological imaging by enabling extraction of high-dimensional embeddings from MRI scans. 
In \cite{yan2023survival} authors propose a Convolutional denoising autoencoder (DAE) network which is combined with a Cox proportional hazard regression loss function to predict survival in the BRATS GBM cohort. In \cite{fu2021survival} authors propose DenseNet for survival classification on the Brats 2018 challenge \cite{menze2014multimodal,bakas2017advancing}. In \cite{gomaa2024comprehensive}, \cite{cerono2024clinical} and \cite{lyu2024survnet} authors proposed transformer, deep neural network  models for survival prediction. 
These models based on convolutional neural nets (CNN) and Vision Transformers (ViTs) capture rich semantic and non-linear patterns; however, they are heavily blackbox in nature. Hence attributing the features used for prognostication can seldom be attributed to tumor regions in GBM. This omission is critical because GBM aggressiveness correlates strongly with topological features such as rim irregularity, tumor periphery, infiltrative outgrowths, and necrotic geometry \cite{d2019pathological,becker2021tumor}. Lack of attributing to these properties in tumor representation make these models clinically unreliable. The other pressing scientific challenge with these existing architectures lie with their susceptibility to scanner variability. Despite internal (training) cohort metrics the models heavily underperform to unseen validation.
\begin{figure}[!ht]
  \centering
  \captionsetup{font=footnotesize}
  \includegraphics[width=0.85\columnwidth,
    trim={5pt 3pt 5pt 3pt},clip]{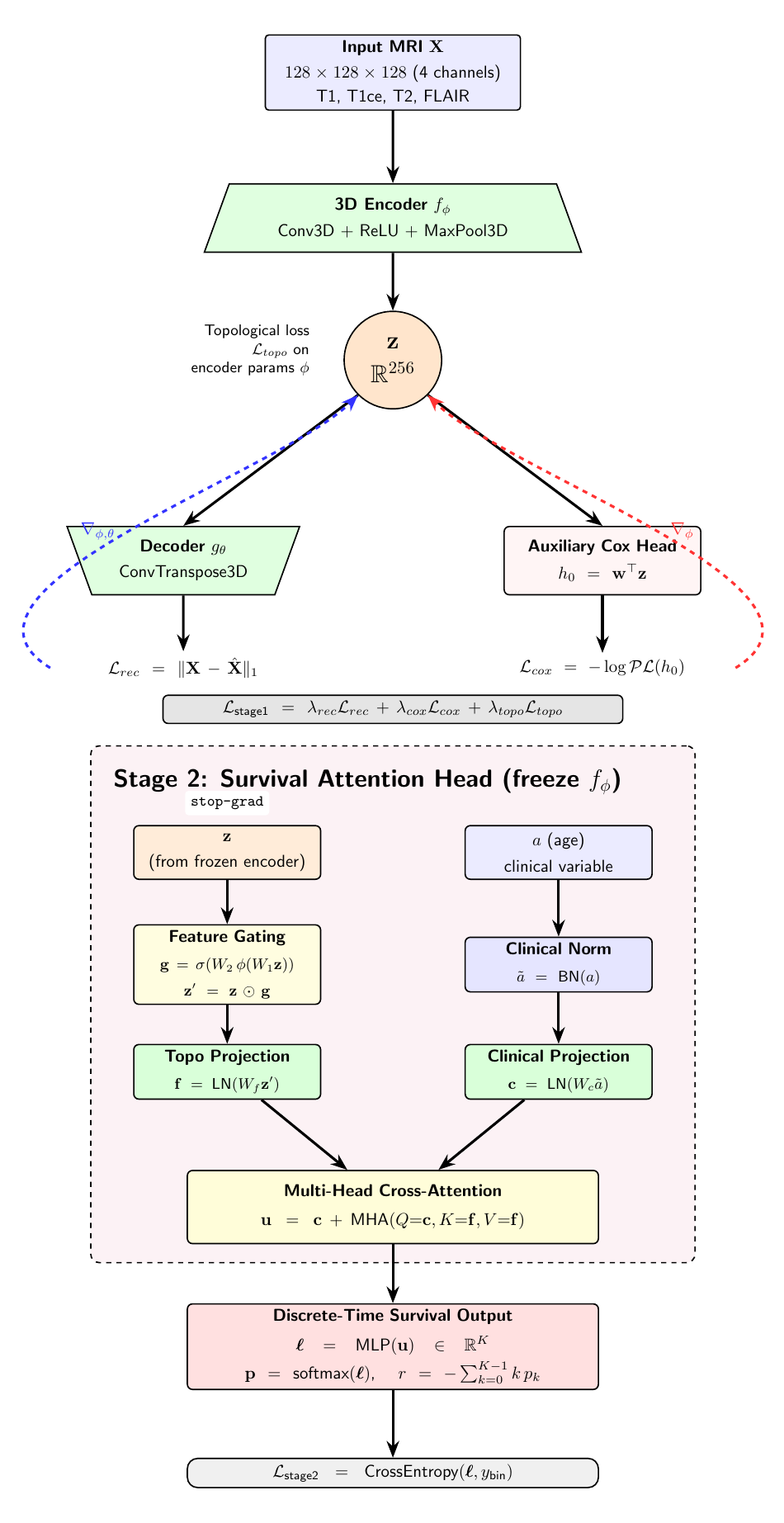}
  \caption{TopoGBM Architectural Framework. The architecture consists of Stage 1: joint representational learning phase which compresses volumetric MRI data into a topologically-regularized latent representation $\mathbf{z}$ and a Stage 2 inference phase that utilizes a cross-attention head to fuse these imaging embedding features with clinical covariate. The final pipeline maps this multimodal representation to discrete-time survival bins to generate an expected risk score $r$.}
  \label{fig:framework}
  \vspace{-10pt}
\end{figure}
To address these gaps, we propose TopoGBM, a GBM tumor heterogeneity representation learning framework trained with neural networks with topology-aware regularization. 
Our approach incorporates a TopoLoss term that constrains the latent embeddings to preserve persistent homology features across multiple filtrations. We explicitly penalize deviations in the Betti-number barcodes, to ensure that critical shape descriptors like connectivity of necrotic cores and the continuity of irregular tumor margins are retained within the latent manifold $\mathbf{z}$. \cite{deb2025toponets}. These topology-constrained latent embeddings from stage 1 are subsequently harmonized across cohorts and passed to a cross-attention survival head in stage 2, where a clinical covariate (age) acts as the query over the imaging embedding to produce discrete-time risk predictions. We re-attribute the latent embeddings to regions across the tumor and normal tissue to explain the hazard and embedding fraction. Our objective is to learn clinically relevant features from MRI using semi-supervised representation learning to make outcome predictions more interpretable and generalizable to external cohorts. 
In multi-institutional experiments across UPENN, RHUH, and UCSF datasets, TopoGBM improves out-of-cohort survival prediction, reduces overfitting, and enhances biological interpretability compared to conventional models. These findings highlight the value of incorporating topological priors into representation learning for robust and clinically meaningful MRI biomarkers in GBM prognosis. We make the following key contributions in this paper: 
\begin{enumerate}[nosep]
\item Brain inspired unsupervised representation learning:
To enforce anatomically meaningful structure in the representation, we introduce a mammalian-brain-inspired topological regularizer (TopoLoss) on an encoder, which encourages topology-preserving organization of learned features, promoting stability to domain shift while retaining clinically relevant morphology. This topology-aware constraint biases the latent space toward robust shape-driven cues rather than spurious texture, strengthening interpretability of downstream risk signals.
\item Multimodal  reconstruction and survival training:
We introduce a multi-task 3D autoencoding framework that learns patient-level embeddings from multi-modal MRI via a reconstruction objective coupled with a survival-oriented hazard constraint. This joint training encourages the latent space to preserve clinically relevant structure while remaining predictive of outcome, producing embeddings suitable for downstream prognostic heads and cross-cohort transfer. We propose a two-stage training strategy: (i) learn generalizable embeddings with the joint model; (ii) freeze the encoder and train  survival head  to explicitly test how much prognostic signal is retained in the learned embeddings.
\item Occlusion-based tumor, peritumor and normal tissue attribution: We provide a region attribution analysis that links the learned representation to tumor spatial anatomy. With occlusion sensitivity, we quantify what fraction of (a) predicted hazard and (b) top embedding dimensions is driven by tumor, peritumoral rings (0–20 vox), and normal tissue, with a stricter decomposition that separates brain tissue from CSF-like regions. This enables TopoGBM to be an interpretable and clinically explainable framework compared to traditional DL models for GBM which are black box in nature.
\item Reconstruction Fidelity Benchmarking: We evaluate reconstruction quality (MAE, MSE, PSNR, SSIM) on training and validation data, and obtain tumor vs non-tumor error decomposition. This demonstrates that the learned representation is not only prognostically useful but also maintains imaging fidelity consistently across datasets, supporting generalization.
\end{enumerate}

\section{Methods}
\label{sec:methods2}
We present \textbf{TopoGBM} as a semi-supervised representational learning framework that (i) learns a compact multimodal latent space representation from 3D MRI via a reconstruction-driven encoder--decoder, (ii) uses the learned embedding (optionally fused with clinical covariates such as age) for survival risk prediction iii) represents the learnt embeddings from tumor and non-tumor regions for clinical relevance instead of being a black box. 
For training TopoGBM we used 3 multicenter cohorts : the University of Pennsylvania Glioblastoma (UPenn) dataset (n=627) \cite{bakas2022university}, the University of California San Francisco Preoperative Diffuse Glioma MRI (UCSF) dataset (n=398) \cite{calabrese2022university} and the Rio Hortega University Hospital Glioblastoma (RHUH) dataset (n=34) \cite{cepeda2023rio}. We split the datasets into a training set (UPENN=492, UCSF=321, RHUH=24) and unseen testing set (UPENN=135, UCSF=77, RHUH=10). For external unseen validation we use the Cancer Genome Atlas glioblastoma cohort (n=97) \cite{bakas2017advancing,bakas2017segmentation}.
Let $i\in\{1,\dots,N\}$ index patients. 
Each patient has four co-registered MRI modalities :
\begin{equation}
X_i = \big[X_i^{(\mathrm{T1})},\,X_i^{(\mathrm{T1ce})},\,X_i^{(\mathrm{T2})},\,X_i^{(\mathrm{FLAIR})}\big].
\end{equation}
Survival outcomes are given by observed time and censoring indicator ($t_i \in \mathbb{R}_{>0}$,$e_i\in\{0,1\}$):
where $e_i=1$ indicates an event (e.g., death) and $e_i=0$ indicates right-censoring.
Optionally, a clinical covariate vector is provided ($a_i \in \mathbb{R}^{p}$). $p=1$in our age-only attention-head implementation.
$\Omega$ the voxel lattice defined as $\Omega=\{1,\dots,D\}\times\{1,\dots,H\}\times\{1,\dots,W\}$.
TopoGBM processes standardized 3D volumes of fixed spatial size. Each NIfTI volume is resampled (via trilinear interpolation) to a common grid size $(D,H,W)$ (\ $128\times128\times128$, raw voxel intensity  $x\in\mathbb{R}$). For each modality volume, we apply intensity min-ax normalization \cite{sinsomboonthong2022performance}.
The resulting input tensor is $\tilde{X}_i \in [0,1]^{4\times D\times H\times W}$.
\subsection{Stage 1: Joint Representation Learning}
\label{sec:stage1}
Here an encoder $f_{\theta}$ is trained to map multimodal MRI volumes to a latent embedding, and a decoder $g_{\phi}$ that reconstructs the input (embedding dimension $d=256$).
\begin{equation}
z_i = f_{\theta}(\tilde{X}_i) \in \mathbb{R}^{d},
\qquad
\hat{X}_i = g_{\phi}(z_i) \in \mathbb{R}^{4\times D\times H\times W},
\end{equation}
The encoder is a sequence of 3D convolution blocks. For layer $\ell$, let $h_i^{(\ell)}$ be the feature map. A block is represented by:
\begin{equation}
u_i^{(\ell)} = \mathrm{Conv3D}\!\left(h_i^{(\ell-1)};W^{(\ell)},b^{(\ell)}\right),
\end{equation}
\begin{equation}
v_i^{(\ell)} = \sigma\!\left(u_i^{(\ell)}\right),
\end{equation}
\begin{equation}
h_i^{(\ell)} = \mathrm{Pool3D}\!\left(v_i^{(\ell)}\right),
\end{equation}
where $\sigma(\cdot)$ is a pointwise nonlinearity (ReLU) and $\mathrm{Pool3D}$ is max-pooling with stride 2.
After $L$ blocks, the output feature map is flattened, given by $\mathrm{flat}_i = \mathrm{Flatten}\!\left(h_i^{(L)}\right)$:
and projected to the latent space by a linear layer $z_i = W_z\,\mathrm{flat}_i + b_z.$
The decoder expands the embedding back to a flattened feature tensor, reshaping, and applying transposed 3D convolutions $\mathrm{flat}'_i = W_d z_i + b_d$, $H_i^{(0)} = \mathrm{Reshape}(\mathrm{flat}'_i)$ followed by upsampling blocks (ConvTranspose3D):
\begin{equation}
H_i^{(m+1)} = \sigma\!\left(\mathrm{ConvTrans3D}(H_i^{(m)})\right),
\end{equation}
ending in a 4-channel reconstruction $\hat{X}_i$.
\begin{figure}[!ht]
  \centering
  \captionsetup{font=footnotesize}
  \includegraphics[width=0.8\columnwidth,
    trim={3pt 3pt 2.5pt 2.5pt},clip]{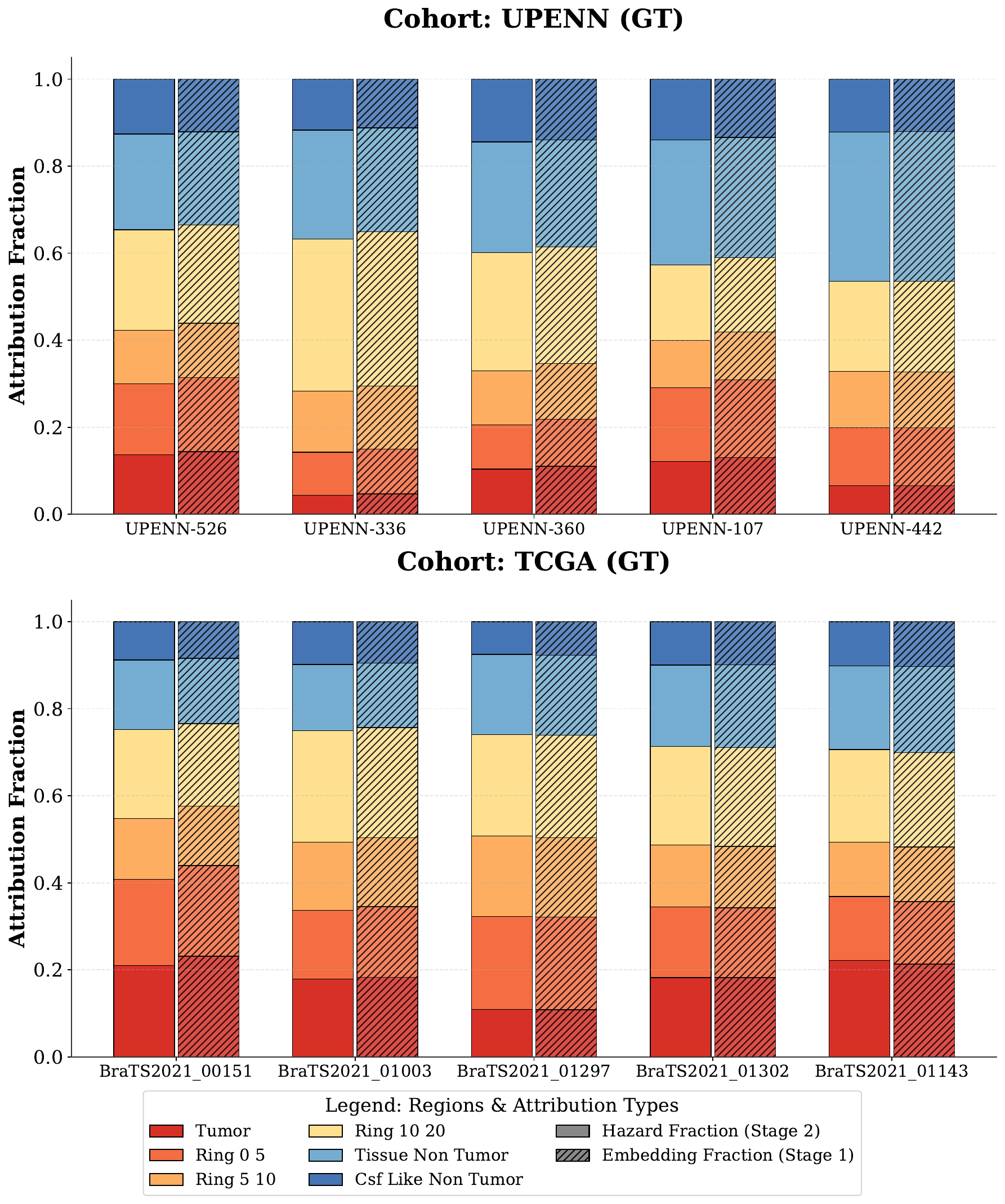}
  \caption{The regional attribution for top-performing exemplars in the expert-verified UPENN and TCGA cohorts, revealing a high degree of spatial alignment between Hazard Fractions (solid, left bars) and Embedding Fractions (hatched, right bars). Nearly 50\% of the model's sensitivity is concentrated within the tumor and immediate peri-tumoral rings (0–20 voxels), demonstrating that the prognostic survival signal is intrinsically anchored to the topologically-encoded tumor morphology rather than healthy anatomical variance.}
  \label{fig:regional_attrib}
  \vspace{-14pt}
\end{figure}
The objective is to minimize voxel reconstruction error using mean-squared error (MSE) across channels and voxels:
\begin{equation}
\mathcal{L}_{\mathrm{recon}}(\theta,\phi)
=
\frac{1}{N}\sum_{i=1}^{N}
\frac{1}{4|\Omega|}
\sum_{c=1}^{4}\sum_{v\in\Omega}
\left(\tilde{X}_i^{(c)}(v) - \hat{X}_i^{(c)}(v)\right)^2.
\end{equation}
We ensure the latent embedding $z_i$ preserves the complex morphology of the tumor we introduce a topological regularization term $\mathcal{L}_{topo}$. We interpret the $d=256$ latent embedding as a $\sqrt{d} \times \sqrt{d}$ ($16 \times 16$) grid $\mathcal{G}$, enforcing structural coherence via a multi-scale filtration:\begin{equation}\mathcal{L}_{topo}(\theta) = \mathbb{E} \left[ \text{dist} \left( \mathcal{D}(\tilde{X}_i), \mathcal{D}(\mathcal{G}_i) \right) \right],\end{equation}where $\mathcal{D}(\cdot)$ represents the persistence diagram derived from the input and latent manifolds. We map $z_i$ to a scalar hazard with a partial likelihood loss based on the Cox Proportional Hazards model ($\mathcal{L}{cox}$ \cite{cox1972regression}), the final joint objective for TopoGBM becomes:\begin{equation}\mathcal{L}{total} = \mathcal{L}{recon} + \tau \mathcal{L}{topo} + \beta \mathcal{L}{cox},\end{equation}
$\tau$ and $\beta$ are hyperparameters modulating the contribution of the structural and prognostic priors.

\subsection{Stage 2: Survival Prediction using Attention}
\label{sec:stage2}
We predict discrete-time survival risk using the learned embedding $z_i$ and a clinical covariate vector $a_i$ (age).

Let $a_i\in\mathbb{R}^{p}$ be clinical covariates (here $p=1$ for age).
The attention head receives \emph{two} inputs $x^{\mathrm{feat}}_i := z_i \in \mathbb{R}^{d},
\qquad
x^{\mathrm{clin}}_i := a_i \in \mathbb{R}^{p}.$:
Thus, age is explicitly used as $x^{\mathrm{clin}}_i$.
Before fusion, a learned feature filter applies a multiplicative gate to the embedding:
\begin{equation}
g_i = \sigma\!\left(W_2\,\sigma(W_1 x^{\mathrm{feat}}_i + b_1) + b_2\right)\in(0,1)^d,
\end{equation}
where $\odot$ is elementwise multiplication, $\tilde{x}^{\mathrm{feat}}_i = x^{\mathrm{feat}}_i \odot g_i,$ and $\sigma(\cdot)$ is the sigmoid.
Clinical variables are batch-normalized and both feature and clinical vectors are projected to a common attention space of dimension $m$:
\begin{equation}
f_i = \mathrm{LN}(W_f \tilde{x}^{\mathrm{feat}}_i + b_f)\in\mathbb{R}^{m},
\end{equation}
\begin{equation}
c_i = \mathrm{LN}(W_c \tilde{x}^{\mathrm{clin}}_i + b_c)\in\mathbb{R}^{m},
\end{equation}
where $\mathrm{LN}$ denotes layer normalization.
To interface with the Multi-head Attention formulation, we treat each modality as a single-token sequence by transposing the projected vectors:
Multi-head attention fuses clinical and imaging embedding information:
\begin{equation}
\mathrm{Attn}(Q,K,V)=\mathrm{softmax}\!\left(\frac{QK^\top}{\sqrt{m}}\right)V.
\end{equation}
In the implemented head, clinical is the query, and embedding features are key/value ($Q_i=C_i, K_i=F_i, V_i=F_i.$)With multi-head attention, $Q,K,V$ are linearly projected per head, attention is computed per head, and then concatenated and projected back to $\mathbb{R}^{m}$. Denoting the attention output by $A_i\in\mathbb{R}^{1\times m}$:
\begin{equation}
A_i = \mathrm{MHA}(Q_i,K_i,V_i).
\end{equation}
A residual ``clinical-centric fusion'' is used ($U_i = C_i + A_i.$)
Finally, $U_i$ is squeezed to a vector $u_i\in\mathbb{R}^{m}$.
The fused vector is mapped to logits over $K$ discrete survival bins ($K=5$), $\ell_i = \mathrm{MLP}(u_i)\in\mathbb{R}^{K}$.
We convert continuous survival time into a discrete bin label $y_i\in\{0,1,\dots,K-1\}$. Let the bin edges be
\begin{equation}
0=\tau_0 < \tau_1 < \cdots < \tau_{K-1} < \tau_K=\infty.
\end{equation}
A standard construction is $y_i = \max\{k:\tau_k \le t_i\}$.
With logits $\ell_i\in\mathbb{R}^{K}$ and class label $y_i$:
\begin{equation}
p_i(k)=\frac{\exp(\ell_i(k))}{\sum_{j=0}^{K-1}\exp(\ell_i(j))},
\qquad
\mathcal{L}_{\mathrm{CE}} = -\frac{1}{N}\sum_{i=1}^{N}\log p_i\!\left(y_i\right).
\end{equation}
To evaluate concordance, logits are converted to a continuous risk score. The expected bin index is $\mathbb{E}[b_i] = \sum_{k=0}^{K-1} k\,p_i(k)$:
In our implementation, earlier bins correspond to earlier death; therefore, higher hazard should correspond to smaller $\mathbb{E}[b_i]$. We define risk as the negated expected bin:
\begin{equation}
r_i = -\mathbb{E}[b_i] = -\sum_{k=0}^{K-1} k\,p_i(k).
\end{equation}
The survival attention head is trained with AdamW($\eta = 5\times10^{-5}$,$\lambda=5\times10^{-2}$):
where $\eta$ is the learning rate and $\lambda$ is weight decay.
\text{Here epochs}$=100$,\quad \text{patience}$=15$.
At each epoch, we compute validation concordance $C_{\mathrm{val}}$ and retain the model with best $C_{\mathrm{val}}$ (early stopping when no improvement beyond a tolerance for 'patience' epochs incorporated).
\textbf{Evaluation Metrics: }
For predicted risk scores $r_i$, concordance index is given by:
\begin{equation}
\mathrm{C}\text{-}\mathrm{index}
=
\Pr\left(r_i > r_j \;\middle|\; t_i < t_j,\; e_i=1\right),
\end{equation}
For each evaluated cohort the trained model outputs $r_i,\quad
\hat{y}_i=\arg\max_k p_i(k),\quad
p_i(0),\dots,p_i(K-1).$:
\begin{figure}[!ht]
  \centering
  \captionsetup{font=footnotesize}
  \includegraphics[width=.80\columnwidth,
    trim={5pt 3pt 5pt 3pt},clip]{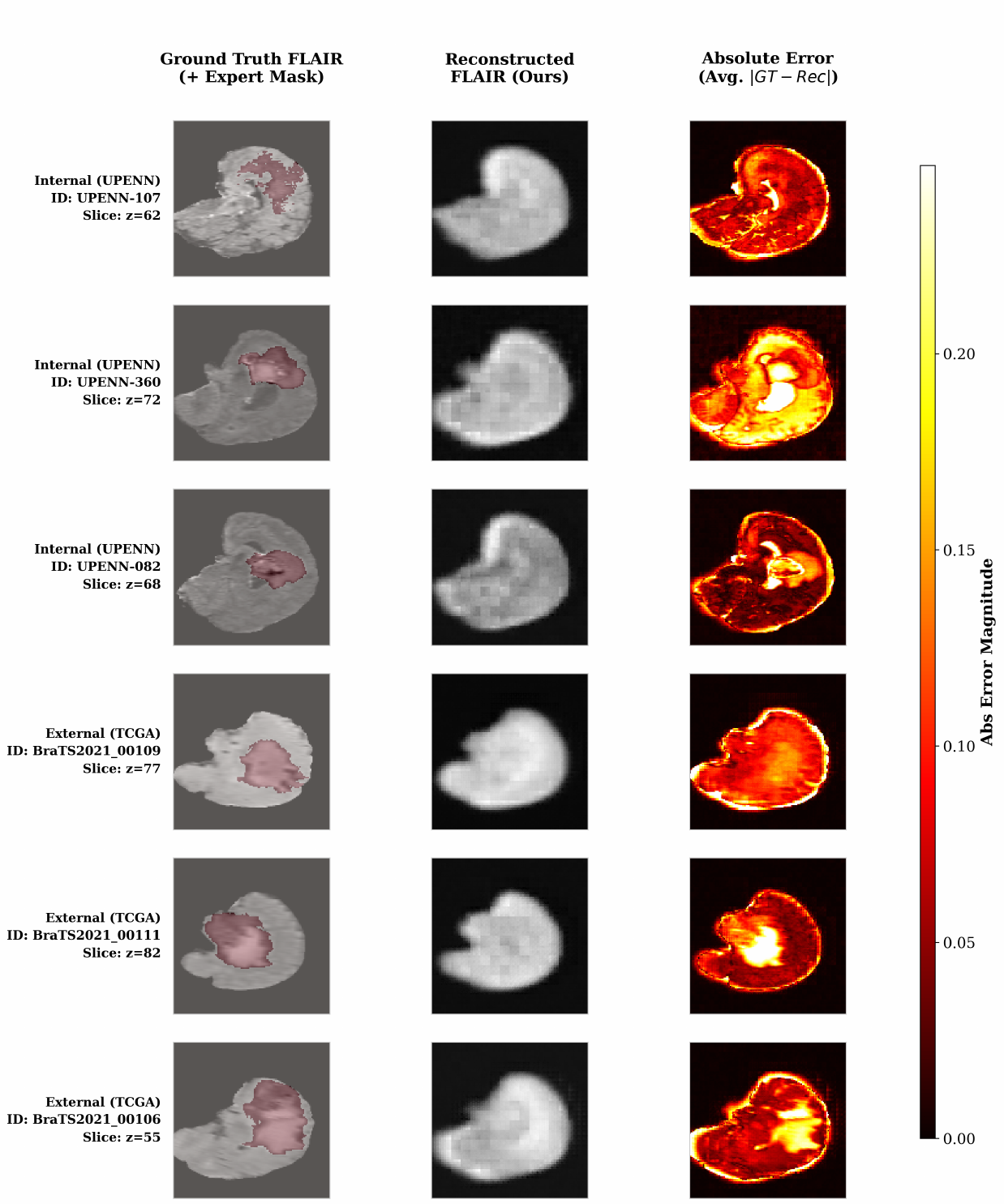}
  \caption{Representative examples from internal UPENN (top 3 rows) and external TCGA/BraTS (bottom 3 rows) showing ground-truth FLAIR with expert tumor mask, reconstructed FLAIR, and the absolute reconstruction error  over the 4 modalities). Across cohorts, the autoencoder preserves global anatomy while residual error concentrates in and around tumor regions, visually supporting tumor-enriched reconstruction discrepancies under domain shift.}
  \label{fig:recon_figure}
\end{figure}
We quantify reconstruction fidelity with mean absolute error (MAE), mean squared error (MSE), peak signal-to-noise ratio (PSNR), and structural similarity (SSIM).
For a modality channel $c$:
\begin{equation}
\mathrm{MAE}^{(c)}_i=\frac{1}{|\Omega|}\sum_{v\in\Omega}\left|X^{(c)}_i(v)-\hat{X}^{(c)}_i(v)\right|,
\end{equation}
\begin{equation}
\mathrm{MSE}^{(c)}_i=\frac{1}{|\Omega|}\sum_{v\in\Omega}\left(X^{(c)}_i(v)-\hat{X}^{(c)}_i(v)\right)^2.
\end{equation}
Overall metrics are averaged across modalities.
With intensity range normalized to $[0,1]$. SSIM is computed on a subset of axial slices:$\mathrm{SSIM}_i = \frac{1}{|\mathcal{Z}|}\sum_{z\in\mathcal{Z}} \mathrm{SSIM}\!\left(X_i(\cdot,z),\,\hat{X}_i(\cdot,z)\right)$, $\mathrm{PSNR}_i = 10\log_{10}\left(\frac{1}{\mathrm{MSE}_i+\epsilon}\right)$
where $\mathcal{Z}$ are slices sampled between 20\% and 80\% of the depth to avoid extreme boundary slices.
Given a tumor mask $M_i\subset\Omega$, we compute region-restricted MAE:
\begin{equation}
\mathrm{MAE}_{i,\mathrm{tissue}}^{(c)}
=
\frac{1}{|M_i|}
\sum_{v\in M_i}
\left|X_i^{(c)}(v)-\hat{X}_i^{(c)}(v)\right|,
\end{equation}
Aggregating across modalities:
\begin{equation}
\mathrm{MAE}_{i,\mathrm{tissue}}=\frac{1}{4}\sum_{c=1}^{4}\mathrm{MAE}_{i,\mathrm{tissue}}^{(c)},
\end{equation}
Here $tissue$ implies for tumor and non-tumor region separately. Our software implementation of TopoGBM will be released on GitHub upon acceptance of the manuscript.

\section{Results}
\label{sec:results}
\begin{center}
\captionof{table}{Comparison of Model C-index and Explainability}
\label{tab:resultscindexexp}
\footnotesize 
\addtolength{\tabcolsep}{-2.5pt}
\begin{tabular}{lccc}
\hline
\textbf{Model} & \textbf{Test} & \textbf{External} & \textbf{Explainable} \\
\hline
DeepSurv\cite{cerono2024clinical}& 0.64 & 0.48 & x \\
Transformer\cite{gomaa2024comprehensive} & 0.36 & 0.51 & x \\
SurvNet\cite{lyu2024survnet} & 0.47 & 0.57 & x \\
\textbf{TopoGBM (Ours)} & \textbf{0.67} & \textbf{0.58} & \checkmark \\
\hline
\end{tabular}
\end{center}
In \ref{tab:resultscindexexp} we observe TopoGBM outperforms existing DL architectures and achieves a C-index of \textbf{0.67} on the internal test set comprising of UPENN, UCSF and RHUH GBM cohorts and C-index of 0.58 on the external validation set comprising of TCGA GBM cohort. Existing DL baselines \cite{cerono2024clinical,gomaa2024comprehensive,lyu2024survnet} show significant performance degradation, especially \cite{agarap2018deep} to 0.48 and \cite{gomaa2024comprehensive} to 0.51, at or below random threshold on the external generalization. This is indicative of TopoGBM's representatio learning with spatially-constrained manifold ensures that prognostic features remain robust across institutional variability and exhibits robust generalization. We also report that TopoGBM is to our knowledge the only DL framework that is explainable compared to existing architectures. 
Nearly 50\% of model sensitivity is concentrated in the tumor and peri-tumoral environment (0--20 voxels). The alignment between hazard and embedding fractions suggests the survival signal is intrinsically tied to the latent geometric organization (\ref{fig:regional_attrib}). 
\begin{table}[t]
\centering
\footnotesize
\caption{Reconstruction metrics by split (mean with 95\% CI)}
\label{tab:recon_split_stats}
\addtolength{\tabcolsep}{-5.5pt}
\begin{tabular}{l c c c c }
\toprule
\textbf{Metric} & \textbf{Train (N=771)} & \textbf{Test (N=209)} & \textbf{Validation (N=97)} \\
\midrule
MAE  & 0.038 [0.037,0.038] & 0.037 [0.037,0.038] & 0.093 [0.088,0.098] \\
MSE  & 0.009 [0.009,0.009] & 0.009 [0.009,0.009] & 0.015 [0.014,0.017]\\
PSNR & 20.491 [20.442,20.541] & 20.522 [20.435,20.611] & 19.076[18.75,19.408]\\
SSIM & 0.657 [0.653,0.661] & 0.659 [0.651,0.666] & 0.754[0.748,0.759] \\
\bottomrule
\end{tabular}
\end{table}
In \ref{fig:recon_figure} across both UPENN and TCGA/BraTS cases, the reconstruction column demonstrates that the model captures large-scale brain structure and intensity layout, indicating stable global reconstruction quality. The error maps show that high-intensity residuals consistently localize to the tumor mask and the peri-tumor localities in both internal and external datasets, while most non-tumor brain tissue exhibits comparatively lower error. This suggests that the compact learned embeddings capture the high-variance pathological components like necrosis, edema, and irregular boundaries. We observe that TopoGBM prioritizes tumor-specific morphology over background artifacts. The general trend of the same tumor error localization in the external TCGA/BraTS rows suggests the effect is cohort independent, contrastive to existing methods in literature.
\begin{figure}[!ht]
  \centering
  \captionsetup{font=footnotesize}
  \includegraphics[width=1.0\columnwidth,
    trim={8pt 8pt 8pt 8pt},clip]{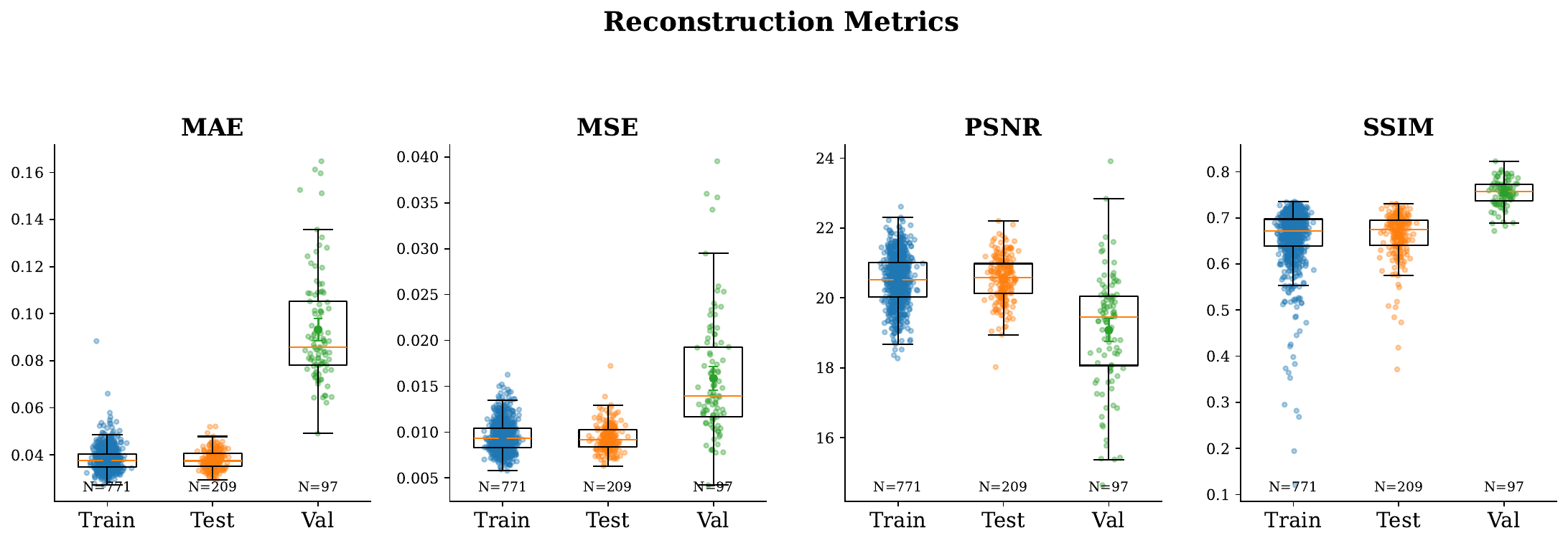}
  \caption{Distribution of per-patient reconstruction quality on Train, Test, and external Val (TCGA) using MAE, MSE, PSNR, and SSIM; each subplot shows a boxplot with points and the sample size (N) for each set}
  \label{fig:recon_metrics_plot}
\end{figure}
TopoGBM demonstrates high fidelity in capturing the 3D morphology of GBM. We observe  a consistently low MAE ($0.038$ and $0.037$) in train and test sets. While external validation dataset shows a slight increase in MAE ($0.093$). We observe an  increase in SSIM: $0.754$) in TCGA cohort, suggesting TopoGBM preserving essential topological and geometric features despite cross-institutional domain shift. GBM being extremely heterogenous with tumor microenvironment irregularities, a moderate PSNR of $20.52$ and $19.07$ in test and validation indicates that the model captures primary tumor signal effectively filtering the stochastic noise and extreme intensity variations within the tumor core. The metrics are also generalizable across cohorts but show much room for further improvement. The topology regularized latent space effectively filters out scanner-specific noise while prioritizing the high-variance structural signatures characteristic of aggressive tumor pathology \ref{fig:recon_metrics_plot}.
\vspace{-10pt}
\section{Conclusion}
\label{sec:conclusion}
Here, we introduced TopoGBM, a novel multimodal framework that integrates spatially-constrained topological embeddings with clinical context for Glioblastoma prognosis. Our results demonstrate that TopoGBM successfully learns scanner-invariant features, evidenced by strongly generalizable C-index, MAE, MSE, SSIM and PSNR. Crucially, we found that nearly 50\% of the model's sensitivity to hazard relies on tumor and peri-tumoral rings asserting clinical evidence of aggressive edema. By utilizing 
patient age in a cross-attention mechanism, TopoGBM achieved a state-of-the-art C-index of 0.67 and 0.58 across unseen samples, significantly outperforming existing DL baselines. While the brain inspired topological regularizer ensures spatial continuity and captures significant tumor specific features, future direction will be into the specific manifold geometries that correlate with treatment resistance.

\vspace{-10pt}
{\small
\bibliographystyle{IEEEbib}
\bibliography{strings}}
    
    
    
\end{document}